\documentclass[a4paper,fleqn]{cas-dc}

\usepackage{cite}
\usepackage{amsmath,amssymb,amsfonts}
\usepackage{algorithmic}
\usepackage{graphicx}
\usepackage{textcomp}
\usepackage{caption}

\usepackage{float}
\usepackage{graphics} % for pdf, bitmapped graphics files
\graphicspath{ {./images/} }
\usepackage{epsfig} % for postscript graphics files
\usepackage{comment}
\usepackage{multirow}
\newcommand{\expnumber}[2]{{#1}\mathrm{e}{#2}}
\newcommand{\pos}{\mathbf{x}}
\newcommand{\etal}{\emph{et al.}}

\usepackage[authoryear,longnamesfirst]{natbib}
\def\tsc#1{\csdef{#1}{\textsc{\lowercase{#1}}\xspace}}
\tsc{WGM}
\tsc{QE}
\tsc{EP}
\tsc{PMS}
\tsc{BEC}
\tsc{DE}
\begin{document}
\let\WriteBookmarks\relax
\def\floatpagepagefraction{1}
\def\textpagefraction{.001}
\shorttitle{Metal Artifacts Reduction via GANs}
\shortauthors{Zihao WANG et~al.}

\title [mode = title]{Inner-ear Augmented Metal Artifact Reduction with Simulation-based 3D Generative Adversarial Networks}                      
\tnotemark[1]

\tnotetext[1]{This work was partially funded by the regional council of
Provence Alpes Côte d’Azur, by the French government through the UCA JEDI
”Investments in the Future” project managed by the National Research Agency
(ANR) with the reference number ANR-15-IDEX-01, and was supported by the
grant AAP Santé 06 2017-260 DGA-DSH.}

% \tnotetext[2]{The second title footnote which is a longer text matter
%   to fill through the whole text width and overflow into
%   another line in the footnotes area of the first page.}

\author[1,2]{Wang Zihao}
\cortext[1]{Corresponding author:}\ead{zihao.wang@inria.fr}
\orcidauthor{https://orcid.org/0000-0001-6534-6641}{Wang Zihao}
%\fntext[fn1]{This is author footnote for second author.}
\author[2,3]{Vandersteen Clair}
\author[4]{Demarcy {Thomas}}
\author[4]{Gnansia {Dan}}
\author[2,5]{Raffaelli Charles}
\author[2,3]{Guevara Nicolas}
\author[1,2]{Delingette Herv\'{e}} %\fnref{fn1}

\address[1]{Université Côte d'Azur, Inria Sophia Antipolis Méditerranée,  2004 Route des Lucioles,  06902 Valbonne, FRANCE}
\address[2]{Université Côte d'Azur, 28 Avenue de Valrose, 06108 Nice, FRANCE}
\address[3]{Head and Neck University Institute, Nice University Hospital, 31 Avenue de Valombrose, 06100 Nice, FRANCE}
\address[5]{Department of Radiology, Nice University Hospital, 31 Avenue de Valombrose, 06100 Nice, FRANCE}
\address[4]{Oticon Medical, 14 Chemin de Saint-Bernard Porte, 06220 Vallauris, FRANCE}

\begin{abstract}
Metal Artifacts creates often  difficulties for a  high quality visual  assessment of post-operative imaging in {c}omputed {t}omography (CT). A vast body of methods have been proposed to tackle this issue, but  {these} methods were designed for regular CT scans and their performance is usually insufficient when imaging  tiny implants. In the context of post-operative high-resolution  {CT} imaging, we propose a 3D  metal {artifact} reduction algorithm based on a generative adversarial neural network. It is based on the simulation of physically realistic CT metal artifacts created by cochlea implant  electrodes on preoperative images. The generated images serve to train a 3D generative adversarial networks for artifacts reduction. %{Depending} on the {method} of {data} collection, our approach can be either supervised or unsupervised, and applied to 3D CT volume artifact reduction.
The proposed approach was assessed qualitatively and quantitatively on clinical conventional and cone beam CT of cochlear implant postoperative images. These   experiments show that the proposed method {outperforms other} general metal artifact reduction approaches.
\end{abstract}

% \includegraphics{figs/grabs.pdf}

% \begin{highlights}
% \item Research highlights item 1
% \item Research highlights item 2
% \item Research highlights item 3
% \end{highlights}

\begin{keywords}
Artifact Reduction, Deep Learning, GAN
\end{keywords}

\maketitle

\section{Introduction}
Computed Tomography (CT) is  one of the most widely used imaging {techniques} in clinical practice.
The physical principles of spiral CT lead to the unavoidable creation of  artifacts in the reconstructed images in the presence of dense materials, {\em i.e{.,}} 
 those composed of atoms with{ }high atomic numbers. {Several} physical phenomena {contribute} to the creation of such artifacts{, including}
X-ray beam hardening, X-ray scatter, electronic noise, edge effects and also the geometrical characteristics of metal parts.
{The} artifacts are commonly found in routine clinical postoperative imaging, for instance due to fixation plates in orthopaedics, cochlear electrode implants in otology, contrast agents, \emph{etc}. These spurious signals in CT images may impair{ postoperative} analysis{.}
For instance, during cochlear implant surgery, an electrode array{ }inserted along the cochlear scala tympani{ }is usually {comprised of a} metal alloy{,} for its high electrical {conductivity}. The existence of metal artifacts in {postoperative} CT makes{ }the evaluation of the {position} of the electrodes along the scala {difficult}. The knowledge of the relative position of the cochlear implant is one of the main {determinants} for assessing the success of the surgery and leads to appropriate and more personalized  patient care.

\begin{figure}[h!]
    \centering
    \includegraphics[width=\columnwidth]{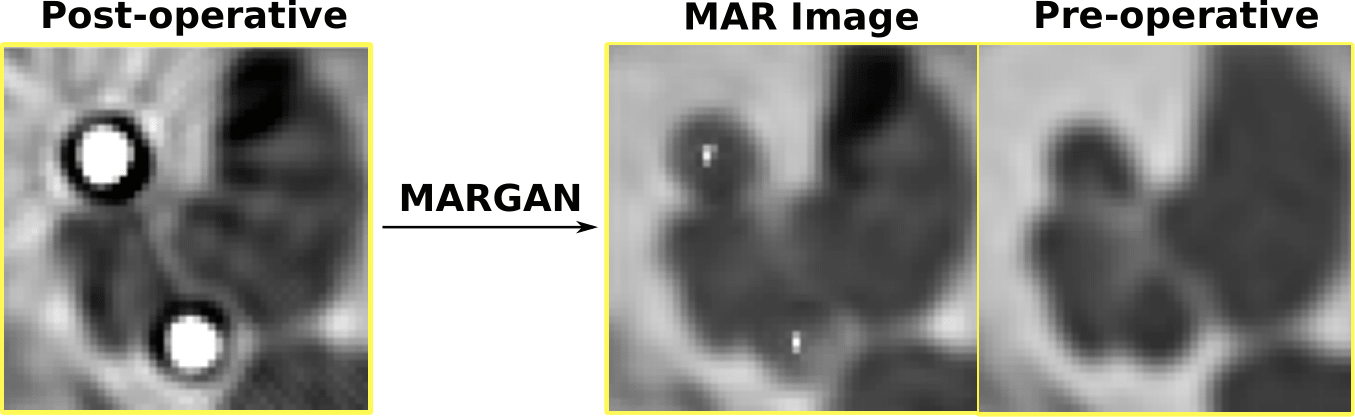}
    \caption{Sketch of the MARGAN algorithm applied on {postoperative} images}
    \label{fig:sketch}
\end{figure}

Metal {artifact} reduction (MAR) methods aim to decrease the extent of {such} artifacts~(see: \ref{fig:sketch}).
Classical non-learning{-}based MAR {algorithms} are divided into two groups: corrupted projection recovery  and iterative image reconstruction{-}based methods~\citep{Mehranian}.
In the former case, projections corrupted by the presence of metal absorbing the X-rays{ }are detected and then replaced by predicted or interpolated values{,} based on prior knowledge ~\citep{Kalender}. The efficiency of the approach is related to the ability to recover the projected signals in the absence of metal parts~\citep{Mehranian}. In the case of iterative methods,
the missing data in image or projection space is estimated {on the basis of} statistical principles{,} possibly including prior knowledge. 
Aside  from Filtered Back Projection (FBP) based methods, \citep{Naranjo} introduced mathematical morphology algorithms for MAR by converting the image {to} polar coordinates centered on the metal artifact.
\subsection{MARGAN approach}
Recently, the field of MAR has been revived by the development {of} deep learning methods that provide supervised mechanisms for extracting relevant image features. 
A number of 2D {Convolutional Neural Network (CNN)}-based MAR methods have been proposed that are summarized in Table~\ref{tab:previous}. \citep{Zhang} introduced {CNNs} as  prior information in the sinogram (projection) space for the inpainting  or sinogram completion task using a  simulated dataset in the training stage. However, this method needs either the original CT sinograms (usually unavailable to the typical user) or to project back the input image in order to fill-in the missing traces. This limits its application in our dataset, and the sinogram-based MAR algorithms tend to generate over-smoothed images due to their filtering effect.

\cite{Huang} developed a deep learning network, \emph{RL-ARCNN}, in image space to predict residual images ({the} difference between {the images with and without artifacts}) to remove metal artifacts in cervical CT images. 
The  2D network  \emph{DestreakNet} was proposed in~\cite{Gjesteby} for streak {artifact} reduction  as a post-processing step in order to recover the details lost after the application of the interpolation-based normalized MAR~\cite{nmar} algorithm. 

\begin{table*}
\centering
\centering
\caption{Summary of major MAR approaches. (In the {dataset collection column}: BH, SC and EN {indicate} Beam{ }Hardening, Scattering and Electronic Noise, respectively)}
\label{tab:previous}
\resizebox{16cm}{!} {
\begin{tabular}{cccccc} 
\hline
                                             & Processing Domain & Inner Ear MAR & 2D/3D & Dataset Collection    & \begin{tabular}[c]{@{}c@{}}\textcolor[rgb]{0.133,0.133,0.133}{Quantitative evaluation}\\\textcolor[rgb]{0.133,0.133,0.133}{on clinical data}\end{tabular}  \\ 
\hline
marBHC                                       & Sinogram          & NO            & 2D    & {Non}-Learning         & YES                                                                                                                                                        \\
marLI                                        & Sinogram          & NO            & 2D    & {Non}-Learning         & YES                                                                                                                                                        \\
NMAR                                         & Sinogram          & NO            & 2D    & {Non}-Learning         & YES                                                                                                                                                        \\
CNN Prior \citep{Zhang}      & Sinogram          & NO            & 2D    & Simulation (BH)       & YES                                                                                                                                                        \\
RL-ARCNN~\citep{Huang}       & Image             & NO            & 2D    & Simulation~(BH)       & YES                                                                                                                                                        \\
DestreakNet~\citep{Gjesteby} & Image             & NO            & 2D    & Simulation~(BH)       & YES                                                                                                                                                        \\
DudoNet++ \citep{Lyu}~       & Sinogram+Image    & NO            & 2D    & Simulation~(BH;SC;EN) & NO                                                                                                                                                         \\
CycleGAN \citep{Nakao}~      & Image             & NO            & 3D    & Unsupervised          & NO                                                                                                                                                         \\
cGAN \citep{Jianing-b}         & Image             & YES           & 2D    & Paired Data           & YES                                                                                                                                                        \\
MARGAN (Proposed)                  & Image             &  YES            &  3D  & Simulation~(BH;SC;EN) & YES                                                                                                                                                        \\
\hline
\end{tabular}}
\label{tab:methods}
\end{table*}

\citep{Lyu} proposed Dudonet++ for 2D CT  metal {artifact} reduction. Their approach relies on processing the {image with artifacts (henceforth referred to as artifact image)} in both sinogram and image spaces in order to restore fine details in the image. 
Their quantitative evaluation shows that the Dudonet++ is effective for {artifact} reduction on simulated CT images but it lacks a quantitative evaluation on {a} clinical dataset. Furthermore, the method  uses {a} beam hardening correction  \citep{Verburg}{,} which is not always optimal{,} for instance in the case of  Cochlear Implant MAR  (see  \ref{tab:evaluation}).

Recently, generative adversarial networks (GAN) were devised for solving MAR problems instead of CNN classification or regression networks, owing to their ability to generate high quality images. 
\citep{Jianing-b} proposed {a} conditional GAN (cGAN) {approach for} CT images with cochlear implants (CI){, using a} collection of paired and registered post{-} and {preoperative} cochlear implant volumes {to train} 2D/3D cGANs for inner ear MAR. A difficulty in this approach is to collect and{, most importantly,} to register ({preoperative}) artifact{-}free {and} ({postoperative}) artifact images. 
This registration problem must be able to cope with the presence of outliers due to the presence of artifacts.

\citeauthor{Nakao} also proposed a MAR method based on CycleGANs for {artifact reduction in} dental filling and neck CT images{.} The approach is unsupervised and  aims to achieve a cross{-}domain (artifact and artifact-free dataset) 
style transformation through {feature} swapping.  This approach does not require {training} on  paired {datasets}, \textit{i.e.{,}} with and without artifacts, but CycleGAN performance significantly {worsens} when unpaired data is used \citep{CycleGAN} for training instead of paired {data}.
This approach was qualitatively compared with the manual corrections available in commercial CT {scans} and quantitatively assessed on synthetic {datasets}. While the output of the CycleGANs seems effective, this method may not be {useful} for the {reduction of} tiny artifacts like cochlear implants{,} due to the difficult separability of artifacts in feature space.

\begin{figure*}[]
\centering
\includegraphics[scale=0.66]{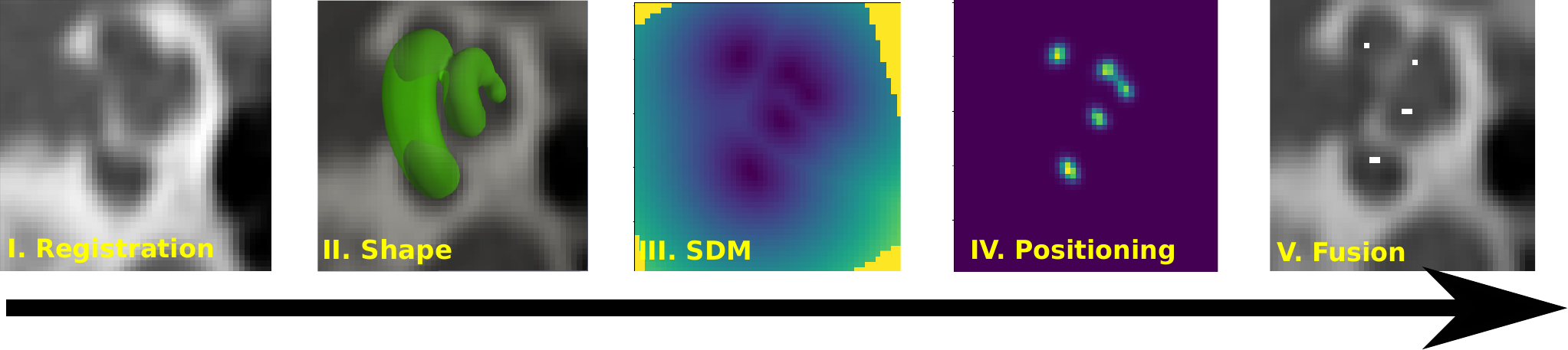}
\caption{Cochlear implant {electrode} positioning simulation; (I) Registration of CT image on a template image; (II) Cochlear shape fitting; (III) Signed distance map generation; (IV) {Electrode} positioning; (V) Image fusion with electrodes.}
\label{fig:CISim}
\end{figure*}
\subsection{Simulating the presence of metal parts}
\label{sec:sim}
In this paper, we propose a GAN-based MAR method {that relies} on simulated training data and {is }suitable for {pre- and postoperative} images. To the best of our knowledge, our approach is  the first MAR algorithm that combines the physical simulation of metal {artifacts} with 3D {GAN} networks.  While classical  GAN-based methods such as \citep{Jianing-b} rely on the existence of paired images with and without artifacts for training, our approach has {several advantages. First,} only {preoperative} images (without artifacts) {are required} for the training stage{, because} the generation of the corresponding artifact image is based on physical simulation. This allows {a} large  set of training images (800 images) {to be used,} without the need for registering {the pre- and postoperative} images. Second, the nature of artifacts can be easily modulated by controlling the complexity of the artifact simulation model complexity. 
Third, we introduce the concept of {
\em augmented metal artifact reduction} by optionally adding landmarks  in the corrected image that indicate the central location  of metal parts. More precisely, in this paper, we show that for the {postoperative} cochlear implant CT images, the  location of each electrode center can be identified 
in the corrected image such that ENT (ear, nose and throat) surgeons can assess the quality of the implantation surgery. Compared to CycleGANs~\citep{Nakao}, the MARGAN  approach allows{ }artifacts {to be easily disentangled} from the background. This is why we believe this approach is probably more appropriate  to attenuate{ }artifacts created by tiny implants. Fourth, MARGAN was evaluated on {postoperative,} cone{ }beam CT images. Finally, MARGAN was developed as a 3D GAN since metal artifacts usually vary continuously between slices. The contributions of the MARGAN framework is summarized in Tab.~\ref{tab:methods}{.}

\begin{figure*}[ht!]
\centering
\includegraphics[width = 2\columnwidth]{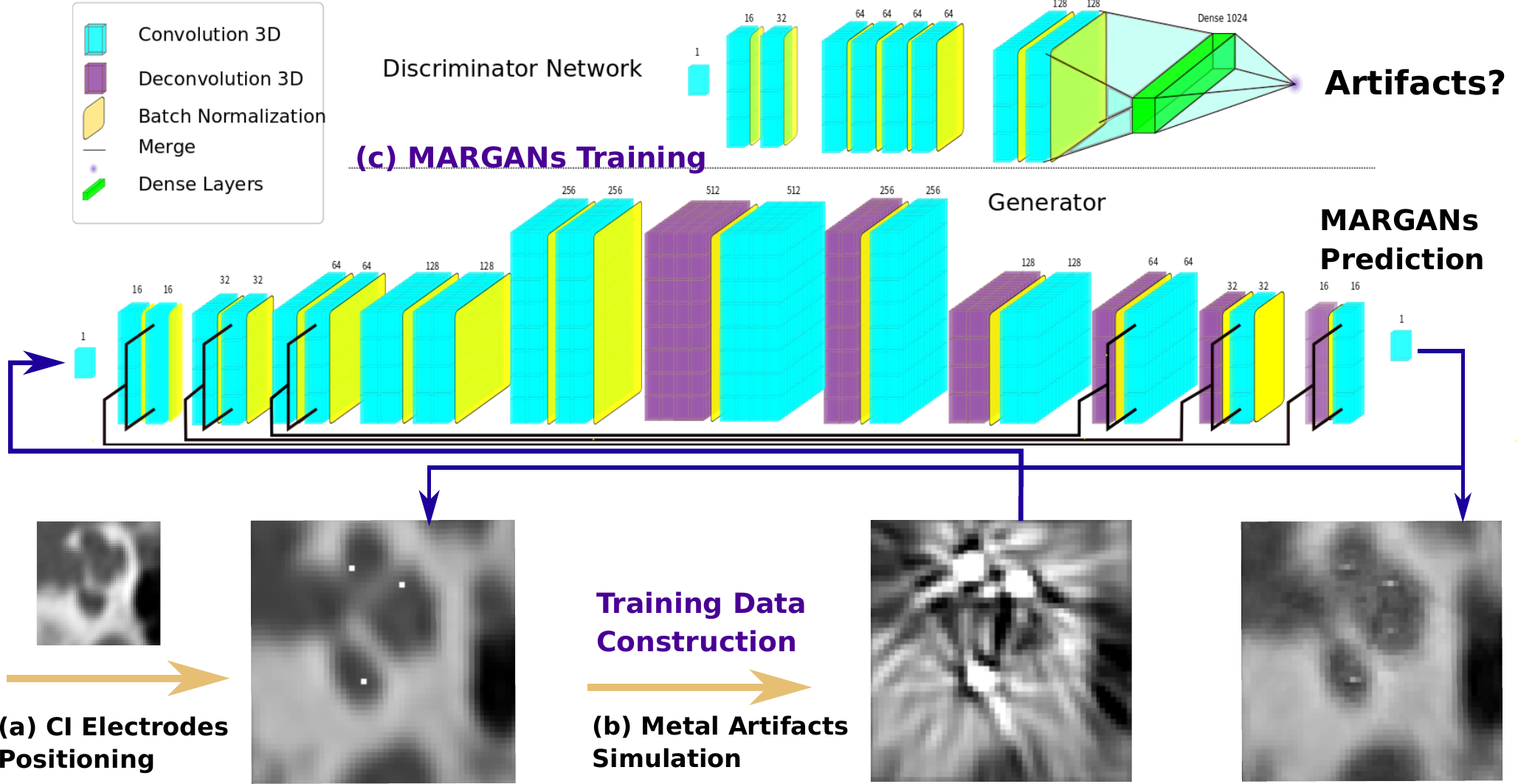}
\centering
\caption{The framework of MARGAN  for metal {artifact} reduction. (a) The cochlear implant positioning simulation; (b) CI metal {artifact} physical simulation. (c) A 3D GAN is trained with simulated and preoperative {datasets}. The generator outputs an image with reduced metal artifacts.
The discriminator network {aims to identify whether or not} the input image includes {metal} artifacts. 
%The generator network accepts an input {artifact} image and {generates a} MAR image.
}
\label{fig:framework_MARGAN}
\end{figure*}

The MARGAN method is based on  two main stages (see Fig.\ref{fig:framework_MARGAN}). In the first stage (Fig.~\ref{fig:workflow}), %CB comment: fix figure references here
given a preoperative image from the  training set, one or several images with metal {artifacts} are generated. This requires a rough segmentation of the structures of interest, the {position} of metal parts (e.g.{,} electrode arrays) and the simulation of artifacts based on {a} CT image formation model. Furthermore, the location of the electrode arrays is added to the generated images.  In the second stage (Fig.~\ref{fig:framework_MARGAN}), a 3D GAN is trained {using} {preoperative} and  corresponding simulated  artifact images. 
The GAN loss is improved by adding a term  based on {Retinex} theory to decrease the image blur in generated images. After training, the GAN is applied on a {postoperative} image without any segmentation or other {preprocessing}. It  results in images with attenuated metal artifacts but also with   landmarks corresponding to electrode centers.

The MARGAN method was applied {to a set of} inner ear CT images to {reduce} the artifacts created by cochlear implants. Qualitative and quantitative results are provided {for} 33 paired {pre- and postoperative CT} images{,} including a comparison with two classical open source MAR algorithms. Qualitative evaluation of {cone beam} CT (CBCT) {postoperative} images is also provided.

This paper extends the initial work published in \citep{Zihao} in several ways. The artifact simulation model is more sophisticated{, }including  scattering effects and electronic noise of the CT system detectors. The algorithm evaluation is more comprehensive{,} with the addition  of  paired CT images, CBCT images and a study of the impact of the {Retinex} loss. 
The {postoperative} electrode position is assessed in {a} few cases with postmortem photographic views of the cochlea. 

The paper is organized as follows: In section \ref{CISIM}, we introduce the CI and CI metal {artifact} simulation procedures (the gray box in Fig.~\ref{fig:framework_MARGAN}). In section \ref{NETWORK}, the network implementation is described (the green box in Fig.~\ref{fig:framework_MARGAN}). Results of the MARGAN algorithm are presented  in section \ref{RESULT}. Sections \ref{DISCUSSION} and \ref{CONCLUSIONS} discuss the contributions  and limitations of the proposed approach.
\section{ Simulation of metal artifacts in CT images}
\label{CISIM}

\begin{figure*}[ht!]
    \centering
    \includegraphics[scale=0.7]{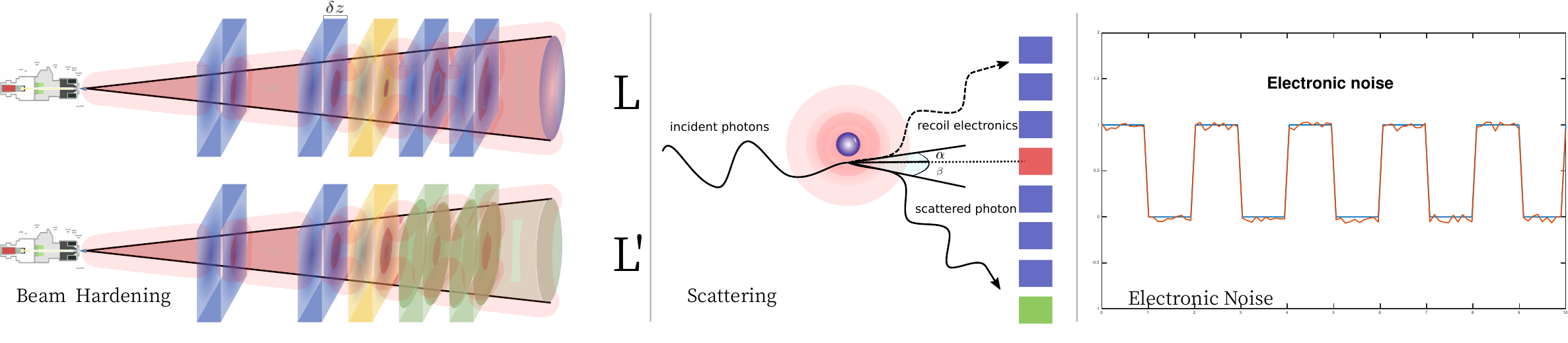}
    \caption{Three main physical{ }effects are considered for simulating metal artifacts. (Left) Beam hardening.   The metal part{, shown} in gold{ }has {nonlinear} X-ray energy absorption{,} thus violating the  Beer-Lambert law. This generates an underestimation  of the material attenuation ratio located after the metal part. 
    %while middle The left above figure shows ideal linear correlation for X-ray attenuation $I$ that the X-ray is still following Beer-Lambert low. However, the realistic phenomena is the metal material absorb the X-ray with non-linear attenuation way which leads the measured $I'\neq I$.
    (Middle) Scattering effect. A scattered photon is abnormally detected by the green detector{, but would} have been detected by the red detector in the absence of scatter. (Right) {The} electronic noise (red) and the corresponding ideal signal (blue).}
    \label{fig:simulatedEffects}
\end{figure*}

The simulation of metal artifacts in CT images from artifact{-}free images {entails} i) {simulating} the presence of metal parts in the images as shown in Fig:~\ref{fig:CISim} and ii) {simulating} the creation of artifacts caused by those metal parts as shown in Fig:~\ref{fig:simulatedEffects}. The former algorithm is completely dependent on the organ{ }or implant considered{,} while the latter is far more generic, based on the physics of image formation.

The processing pipeline to generate the training set for the MARGANs is displayed in Fig.~\ref{fig:CISim}. In this section, we consider the case of {preoperative} CT images of the inner ear prior to{ }cochlear implant surgery. The objective is therefore to simulate{,} in {these} preoperative {images}, the addition of metal electrode arrays associated with the implant.

The 3D CT volumes of the inner ear{,} written as $I(\pos)$, are first rigidly registered on a template image by a block matching algorithm~\citep{Ourselin}. The template is a sample CT image that has been manually cropped around the temporal bone. The registration is necessary to cope with the {variations} of field of {view} and pose {in} the input image dataset. 
%image acquisitions  This alignment is required in order to use the cochlea shape fitting model to fit the cochlea shape. 
A region of interest (ROI) is then cropped to get a cochlear  volume suitable for further processing. We then fit a parametric  shape model~\citep{Demarcy2017} %CB comment: a parametric model of the cochlea shape? or simply a parametric model? (because shape of the cochlea is in next part of sentence)
to automatically reconstruct the shape of the cochlea (step (II) of Fig:~\ref{fig:CISim}).
The {accuracy} required for {the} registration and segmentation steps is limited. 

The signed distance map~\citep{midl2020} from the fitted triangular mesh of the parametric shape model is generated as  shown in step (III). It is then thresholded (step (IV)) to create a 3D tubular binary mask near the center-line of the scala tympani of the cochlea. This mask corresponds to the probable location of the electrodes after a CI intervention. Finally, in step (V), the voxel values in Hounsfield {units} (HU) of the mask region  are then set to $3071HU$ which is the maximum detectable HU of the CI metal artifacts. This {creates} the image $I^{\mathrm{train}}(\pos)$ used for training the GAN network.

\subsection{Simulation of beam hardening, scattering and electronic noise due to metal parts}
\label{subsec;simulation}
The metal parts have large absorption ratios of X-ray energy which is {the cause of}  the visible artifacts in CT images. It impacts the whole image formation process {through} several physical effects. Our {previous} work~\citep{Zihao} only {considered} the simulation of {the} beam hardening effect{,} inspired by the work of \citep{Zhang}. In this paper, we improve the realism of the simulated artifacts by also including the X-ray scatter effect through Monte{ }Carlo simulation and the detector electronic noise{.} The three main physical effects governing the generation of metal artifacts are described below{, along with} the processing pipeline.

\paragraph{Beam hardening effect}
For a monoenergetic X-ray source entering a material of thickness $ \delta z$ along direction $z$ at position $x,y$, the number of photons  $L(x, y, \delta z)$ is given by the  Beer-Lambert law : $L(x, y, \delta z) = L_0 e^{-\mu(x, y) \delta z}$ where $L_0 $  is the initial photon number and $\mu(x, y)$ is  {the} linear attenuation coefficient of the material.

The attenuation coefficient depends on the energy of the input photon $\mu(E_v)$, and therefore for a polychromatic X-ray beam having 
the energy distribution (or spectrum){,} $\phi(E_v)$,
the  number of photons received by the entire detector surface is then:
\begin{equation}
L =  \intop\nolimits_{E_0}^{E_n} (\phi(E_v)  e^{ -\iiint \mu(x, y, z, E_v) dxdydz}+S(E_v))  dE_v
\label{eq:photoNum}
\end{equation}

where $E_0$ and $E_n$ are the minimum and maximum energies for a fixed tube peak voltage, and { } $S(E_v)$ is {an} additive offset that captures X-ray scattering.  

\paragraph{Scattering effect} The Compton effect applies to incoming {X}-ray photons that interact with the free electrons {in} the traversed materials. This effect  results in random changes {(scatter) in the directions} of the photons, {which} may still reach the detector plate despite collimator devices. The Compton scatter is enhanced in the presence of metal parts, thus resulting in an offset in the number of photons $S(E_v)$  and leading to a reduction {in the} image contrast. Computing this additional scatter  is very complex as it depends on the projected plane {and} the material and geometry of the tissue surrounding the metal parts. To this end, we use Monte{ C}arlo simulation to estimate the offset value $S(E_v)$ for different detector positions and {orientations}. The governing equation for the simulation provides the emission energy $E_{p}(\beta)$ of a polychromatic ray deviating {by} an angle $\beta$ from its initial trajectory :
\begin{equation}
E_{p}(\beta) =  \int \frac{E_v}{(1+E_v/m_ec^2)(1 - cos(\beta))}dE_v
\label{eq:comptonEng}
\end{equation} where $m_e$ is the electron mass and $c$ the speed of light. To estimate the scatter effect inside the cochlea on X-ray detectors, we use the \citep{zubal} head phantom where metal parts are roughly positioned inside the temporal bone. Based on the MCGPU software~\citep{Badal} performing GPU  Monte Carlo simulations of photon transport in voxelized geometry, we simulate thousands of X-ray photon trajectories at different energies, positions and orientations through the head and produce both the scatter-free sinogram $F(E_v)$ and the scatter sinogram offset $\tilde{S}(E_v)$. The scatter sinogram offset is corrected by a scale factor such that the resulting  \textit{scatter to primary ratio} $\alpha=\frac{\mathrm{mean}(\tilde{S}(E_v))}{\mathrm{mean}(F(E_v))}$, falls within the range of 0.1\% to 2\%, which was experimentally found by Glover \etal~\citep{Glover}. This is simply done by randomly picking a ratio $\alpha_r$ within 0.1\% to 2\% and computing \begin{equation}
S(E_v)=\frac{\mathrm{mean}(F(E_v))}{\mathrm{mean}(\tilde{S}(E_v))} \alpha_r~\tilde{S}(E_v)
\label{eq;normalizationScatter}
\end{equation} The {same} ratio $\alpha_r$ is {used} for simulating all sinograms of the same image to obtain spatially consistent artifacts. 

The computation of the scatter offset is  dependent on the {X}-ray energy, position, and orientation but is independent of the input image as it relies on the digital head phantom augmented with metal parts next to the temporal bone. Only the {scatter to primary ratio} varies between different volumes. This implies that the scatter sinograms can be precomputed{,} thus alleviating the {computational} load {when generating} images with metal artifacts.

\paragraph{Detector Noise} Once photons hit the x-ray detector, the scintillator transforms the deposited energy into visible light, while a photomultiplier translates this light into an electric signal. In this process, some electronic noise is introduced which can be modeled by  a zero mean Gaussian distribution~\citep{NsimBenson} {with} standard deviation $\sigma_e$: $N(0, \sigma^2)$. The signal measured  in  each sinogram $L_{\mathrm{final}}$ can then be written as:
$L_{\mathrm{final}} = L + \mathcal{N}(0, \sigma_e^2)$
 where ${L}$ is the energy deposited as described Eq.{ }\ref{eq:photoNum} and $\sigma_{e}^2 = 0.04$.
 
 \begin{figure*}[]
\centering
\includegraphics[width =2\columnwidth]{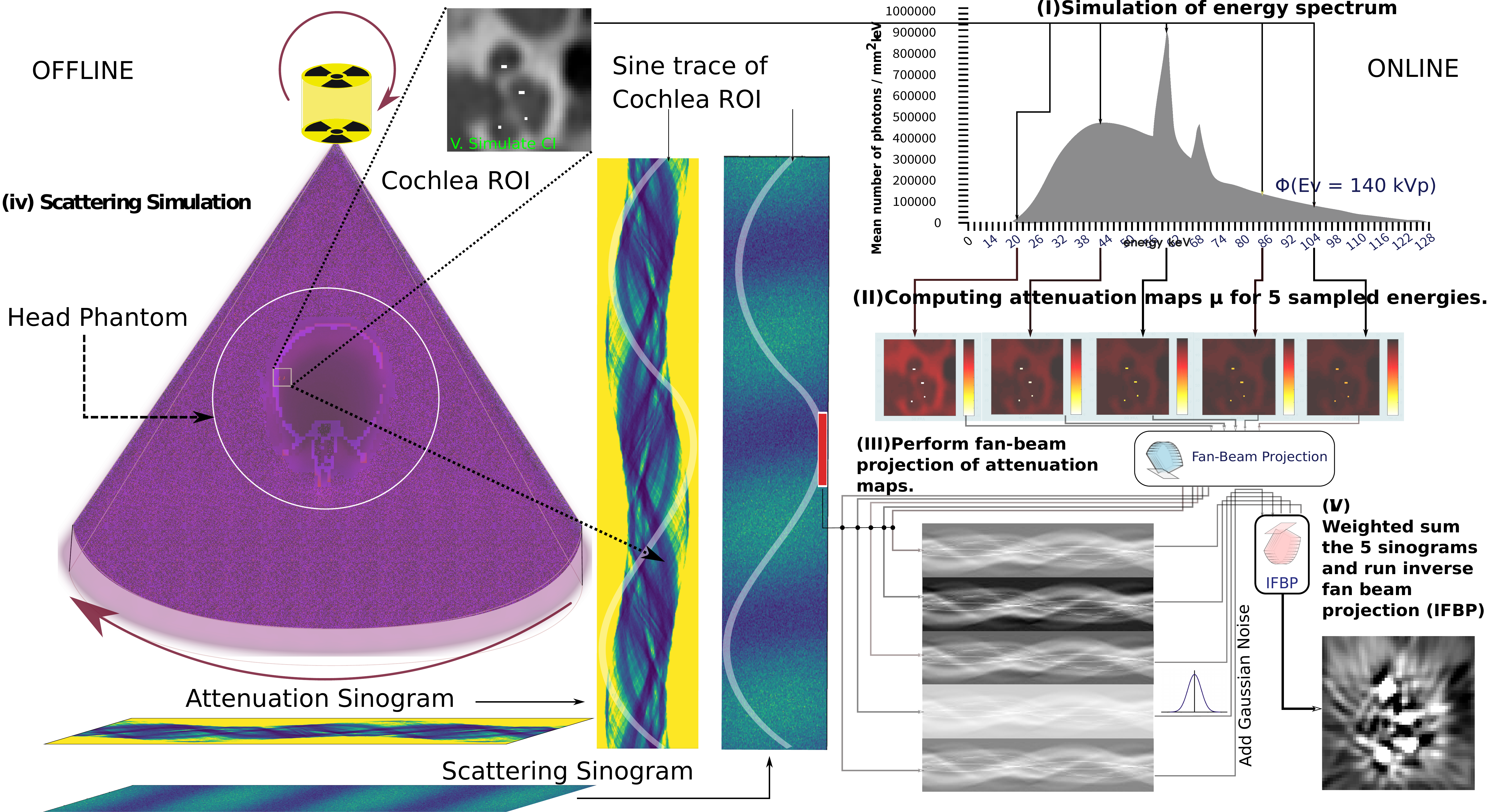}
\caption{Pipeline of metal {artifact} simulation.
%starting from  and ending with the simulated post-operative image. 
Given a {preoperative} image with simulated implants, the simulation starts from the computation of attenuation maps (steps I -II) %CB comment: remove nested parentheses?
for {the} cochlea ROI volume based on the energy spectrum of {the} X-ray tube. {Step} (III) performs fan-beam projection to simulate the sinograms of the attenuation map. {(IV)} Monte{ }Carlo simulation of scattering effects is performed offline on a head phantom for the generation of the scattering sinograms whose traces  {in} the  ROI are randomly chosen, then normalized and added to the combined attenuation map sinograms. 
(V) Gaussian {electronic} noise is added and then  inverse fan{-}beam projection is performed to get the final simulated {artifact} images.}
\label{fig:workflow}
\end{figure*}

 \paragraph{Simulation pipeline} The overall metal artifact simulation pipeline is described in Fig.{ }\ref{fig:workflow}. In the first step, we use an X-ray energy spectrum $\phi(E_v)$  extracted from a CT manufacturer dedicated site~\footnote{https://www.oem-xray-components.siemens.com/x-ray-spectra-simulation} for a tungsten anode tube at 140 $kVp$.  The spectrum is sampled at  five  energies from which attenuation maps $\mu(x, y, z, E_{v_i})$ are generated. This computation is based on the Hounsfield unit formula and the water absorption coefficients as a function of energy. We then perform fan-beam projection (Step III) of the {five} attenuation maps to produce {sinogram}-like images representing absorbed energy on {the} CT detectors. The scattering and attenuation sinograms are precomputed on a head phantom for various orientations and positions of the source. The projection of the ROI of the head where metal parts have been inserted creates a sine trace on the scattering and attenuation sinograms. This trace is randomly sampled, then normalized as in Eq.{ }\ref{eq;normalizationScatter} to obtain a plausible scatter to primary ratio. It is then added to the electronic noise and to the weighted sum of the {five} sinograms (Step IV) and { }a discretization of Eq.~\ref{eq:photoNum}. %CB comment: perhaps check this change is correct - they added a discretized form of Eq. 1, which represents the attenuation map
 Finally, inverse fan{-}beam projection produces the output image with metallic artifacts (Step V).
 
\begin{figure}[!htbp]
\centering
  {\includegraphics[width=\columnwidth]{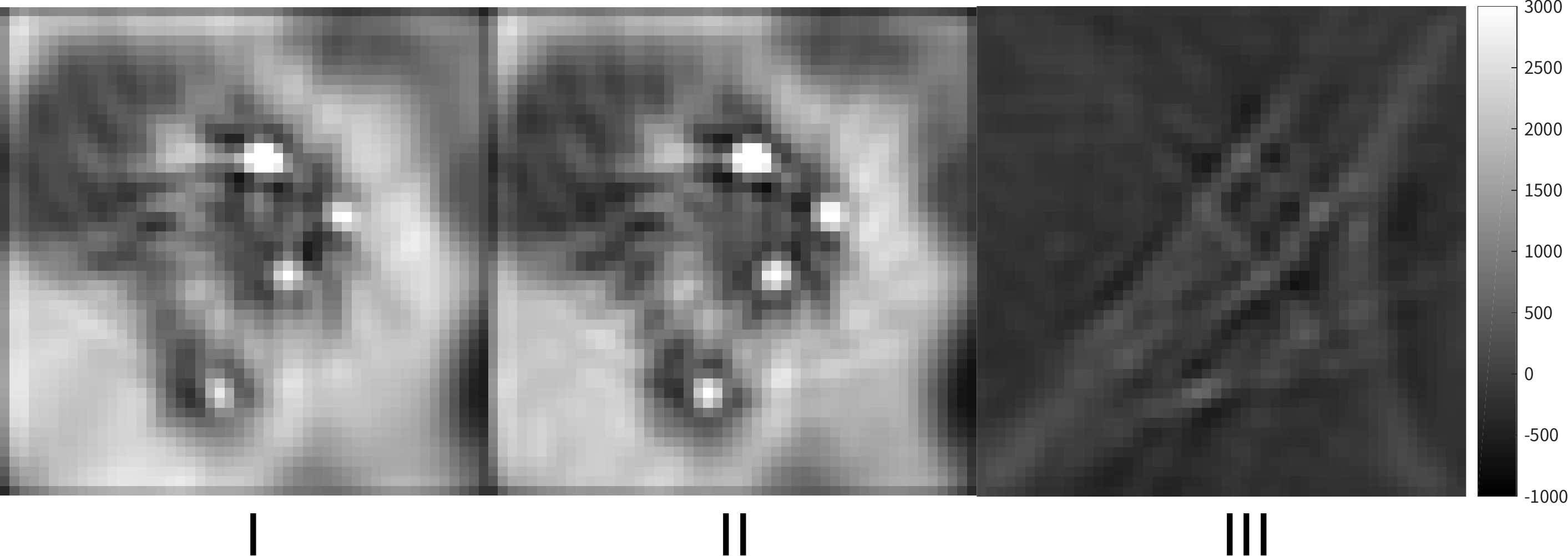}}
  \caption{Noise  and beam hardening images. (I) Simulation with scattering effect and electrical noise. (II) Simulation {with} only beam hardening. (III) The subtraction map between the two simulations.}
  \label{fig:scatteringImage}
\end{figure}

 The difference between simulated images with and without scattering noise is shown in Fig. \ref{fig:scatteringImage} with a subtraction map. We see that scattering and electronic noise can introduce significant new artifacts.
\section{GAN{-}based Metal {Artifact} Reduction}
\label{NETWORK}
Given pairs of {preoperative} and simulated {postoperative} images, we aim to train a network that generates the former given the latter as a way to reduce metal {artifacts}. The use of a GAN to tackle the MAR issue  is motivated by the successful {use} of 2D and 3D GANs such as SRGAN~\citep{Ledig,Sanchez2018} to solve  imaging Super-Resolution (SR) problems. 

\subsection{Network Overview} 
In MARGAN, two neural networks are used{:} the generator network produces the MAR images {and} the discriminator network indicates whether the input image contains metal artifacts or not. 
The generator network{,} ${G}_{w_g}${, with} weights{,} $w_g${,} aims at modeling the mapping between the  image with artifacts{,}  $I^{\mathrm{m}}$, and the simulated artifact-free image{,}  $I^{\mathrm{train}}$. We {denote by}   $I^{MAR}$ the 3D image {created} by the generator network{,} which should be as close as possible {to} $I^{\mathrm{train}}$. The discriminator neural network{,} ${D}_{w_d}${,} tries to detect the presence of artifacts in    the generated MAR {images, }  $I^{MAR}$. To train ${G}_{w_g}$ and ${D}_{w_d}$ networks,   the sum of discriminator and generator losses is optimized as detailed below.

\subsection{Network Architecture}
The generator network architecture %vastly differs from \citep{Sanchez2018} as it
is similar to U-Net with convolution and deconvolution layers, skip connections and batch normalization layers to improve the training efficiency (see Fig~\ref{fig:framework_MARGAN}). Moreover, unlike \citep{Sanchez2018} which is patch based, the input to
the network consists of  full 3D images as it is compatible with GPU memory. 
The number of filters increases gradually from 1 to 512,  a number of feature maps {that can} %CB check this change is correct, in case I have misunderstood - original was "... feature maps, to fit on a 11Gb..."
fit on {an} 11{ }Gb video-memory GPU card. %When the filters number begin to decrease, after each convolution layer a deconvolution layer is applied, the deconvolution layer will reconstruct the feature maps to target size gradually. In the last layer, convolutional layer produces the final generated MAR image. 
The discriminator network follows that of \citep{Sanchez2018} with eight groups of convolution layers and batch normalization layers combined sequentially.

\subsection{Loss Functions}
\paragraph{Discriminator Loss}
The discriminator  network{,} ${D}_{w_d}${,} is trained {using} output images from the generator network{,} $I^{MAR}=G_{w_g}(I^{\mathrm{m}})${,} and{ }images without any metal artifacts{,} $I^{\mathrm{nm}}$. Following
\citep{Sanchez2018}, the discriminator loss enforces the ability of the discriminator network to {distinguish} the artifact{-}free images{,} $I^{\mathrm{nm}}${,} from the generated ones{,} $I^{MAR}$ :
\begin{align*}
\mathop{\arg\max}_{w_d}~L_{D} &= \mathbb{E}_{x\sim I^{\mathrm{nm}}}\log\left(D_{w_d}(x)\right) +\\ &\mathbb{E}_{y\sim I^{\mathrm{m}}}
\log\left(1-D_{w_d}(G_{w_g}(y))\right)
\end{align*}

\paragraph{Generator Loss} The objective of the generator network is to produce an image{,} $I^{MAR}=G_{w_g}(I^{\mathrm{m}})${,}  as close as possible {to} the target image{,} $I^{\mathrm{train}}$. This is why the first loss term is the mean square error (MSE), $L_{mse} = \mathbb{E}_{y\sim I^{\mathrm{m}} }(|I^{\mathrm{train}} - G_{w_g}(y)|^2)${,} 
%and the Absolute Difference Error (ABS)  $L_{abs}$ 
to encourage {a} similarity between generated and {target} voxels. 
But using only the MSE loss leads to blurred MAR images with a lack image detail at high frequencies. 
To avoid this excessive smoothing, we 
propose a new loss term based on Retinex theory~\citep{Edwin}. This theory is mostly used to improve images seriously affected by environmental illumination. The Retinex theory assumes that a given image can be considered as the product of environmental brightness (or illumination){,} $L(x, y)${,} and the object reflectance{,} $R(x, y)$. This reflectance map contains high frequency details and is unaffected by the illumination condition, a property referred to as the  {c}olor {c}onstancy {p}henomenon. The objective of Retinex{-}based algorithms is to recover the reflectance image from the original one. In single{-}scale {R}etinex approaches~\citep{Zhang2}, the environmental brightness is simply a Gaussian blur version of the input image %CB is simply the input image blurred by a Gaussian function?
and therefore $\log(R(x,y))=\log (I(x,y))- \log (I(x,y)*{\mathcal{N}}(0,\sigma))$ where ${\mathcal{N}}(0,\sigma)$ is a Gaussian function of standard deviation $\sigma$, and $*$ is the convolution operator.
This leads us to introduce the following Retinex loss to make its illumination part as close to 1 as possible :
\begin{equation}
L_{retinex} = \mathbb{E}_{Y\sim I^{\mathrm{m}}}\frac{|G_{w_g}(Y) - 
e^{\log{G_{w_g}(Y)}- \log{G_{w_g}(Y) *{\mathcal{N}}(0,\sigma)}}|}{|Y|}
\end{equation}
where the expectation is taken over the image domain. 
This loss definition ensures numerically stable evaluations and
enforces salient features in the image that would {otherwise} be attenuated{.} {Combining it} with the adversarial term $L_{adv}=\frac{1}{2}|D_{w_d}({G_{w_g}})-1|^2$ as in \citep{Sanchez2018},  the full optimization target of the generator is:
\begin{equation}
\mathop{\arg\min}_{w_g}~ L_{generator} = \alpha \cdot L_{retinex} + L_{mse} + L_{adv} 
\label{Eq: GenLoss}
\end{equation}
where $\alpha$ is a parameter controlling the influence of the Retinex loss.

\begin{table*}
\centering
\caption{Material Mapping Table for Voxel Conversion to MCGPU File}
\begin{tabular}{llllllll}
                & air      & water & bones & muscle & titanium & soft tissue & fat    \\ 
\hline
MC-GPU MATERIAL & 1        & 15    & 4     & 2      & 16       & 3           & 6      \\
DENSITY [g/cm$^3$] & 0.001205 & 1.000 & 1.990 & 1.041  & 4.506    & 1.038       & 0.916  \\
\hline
\end{tabular}
\label{tab:material}
\end{table*}

\section{Results}
\label{RESULT}
\subsection{Dataset}
\subsubsection{Training data}
The cochlea dataset was collected from the Radiology Department of the Nice University Hospital with a GE LightSpeed CT scanner without any metal {artifact} reduction filters. The preoperative dataset includes 1000 temporal {bone} images (493 left and 597 right) from
597 patients. The original CT volumes are registered to a sample image by a pyramidal block-matching 
algorithm in order to spatially normalize all images{, then they are resampled} with 
$0.2 \times 0.2 \times 0.2 mm^3$ voxel size. They were then cropped {to volumes of} $60 \times 50 \times 50$ {voxels} around the cochlea region. %CB check this change is correct - original was "cropped as 60x50x50 volumes around..."
We then simulated on all volumes, the insertion of CI electrodes and the generation of metal artifacts as described  in section~\ref{CISIM}. This {created} a set of  1000 pairs of images,  with and without metal artifacts.

\subsubsection{Evaluation Data}
The evaluation dataset \#1 includes 33 cadaver temporal {bone} CT images collected from the same site from different bodies. The imaging protocol {was} the same as for the training dataset but was performed before and after the implantation of CI, thus leading to 33 pre{-} and {postoperative} image pairs. The temporal bones were ground
by an{ }ENT (ear, nose and throat)
surgeon, approximately along a plane perpendicular to the cochlear modiolar
axis at the bottom of the scala tympani as shown in Fig.~\ref{fig:electrodeCompare}. %CB check order of figures - they don't seem to be referred to in the order they appear
Pictures of the ground bones were acquired in order to visualize the electrode array.

Finally, the second evaluation dataset includes  8 {postoperative} images that were acquired on a Carestream 9600  cone beam CT (CBCT) following the CI surgery. {These} images  were resampled, registered and cropped following the same processing pipeline as the training set.   %implantation of  systemimages fro,  scanner images collected from 8 CI 
%o After the collection of CT scanning pairs, grinding photography positioning to the cochlea implanting post-operative was taken . 
\begin{table}[!htbp]
\centering
\caption{Dataset Summary: {Preoperative and postoperative} {refer to images} collected before and after Cochlear Implant, respectively.}
\begin{tabular}{llll}
Dataset     & Pre-Op & Post-Op &  Photography  \\ 
\hline
Training   & 800           & 0              & 0                     \\
Validation & 200           & 0              & 0                     \\
Evaluation  CT& 33            & 33             & 33                    \\
Evaluation  CBCT& 0            & 8             & 0                    \\

\hline
\end{tabular}
\label{tab:conversion}
\end{table}

\subsection{Implementation details}

\paragraph{{Artifact} simulation}
A polychromatic X-{r}ay source was simulated with MC-GPU v1.3, a GPU{-}based Monte Carlo {simulator} of photon transport in voxelized geometry~\citep{Badal}. To simulate the scatter effects, we simplified the contents of the human head by assuming it consists of  air, water, soft tissue, bone, muscle and unalloyed titanium. Cochlear CT voxel values were converted to MC-GPU v1.3 units based on the material mapping {in Table} \ref{tab:material}.
The simulation of scatter was performed offline  on a GPU parallel computing cluster.   The beam{ }hardening maps and the final simulation volumes were computed  with Matlab 2017a on a Dell Mobile Workstation with Intel(R) Core(TM) i7-7820HQ @ 2.90GHz CPU.

\paragraph{Neural Networks}
The networks were trained with a RMSprop optimizer~\citep{Arjovsky} with learning rate $l_rg = \expnumber{1}{-4}$ for the generator and $l_rd = \expnumber{1}{-3}$ for the discriminator. The MARGAN was implemented with Tensorflow and the weight of Retinex loss was set to $\alpha = \expnumber{5}{-5}$. The batch size and number of epochs were set as $1$ and $20$ respectively. 

\subsection{Clinical Evaluation}
\begin{figure}[]
\centering
\includegraphics[width=\columnwidth]{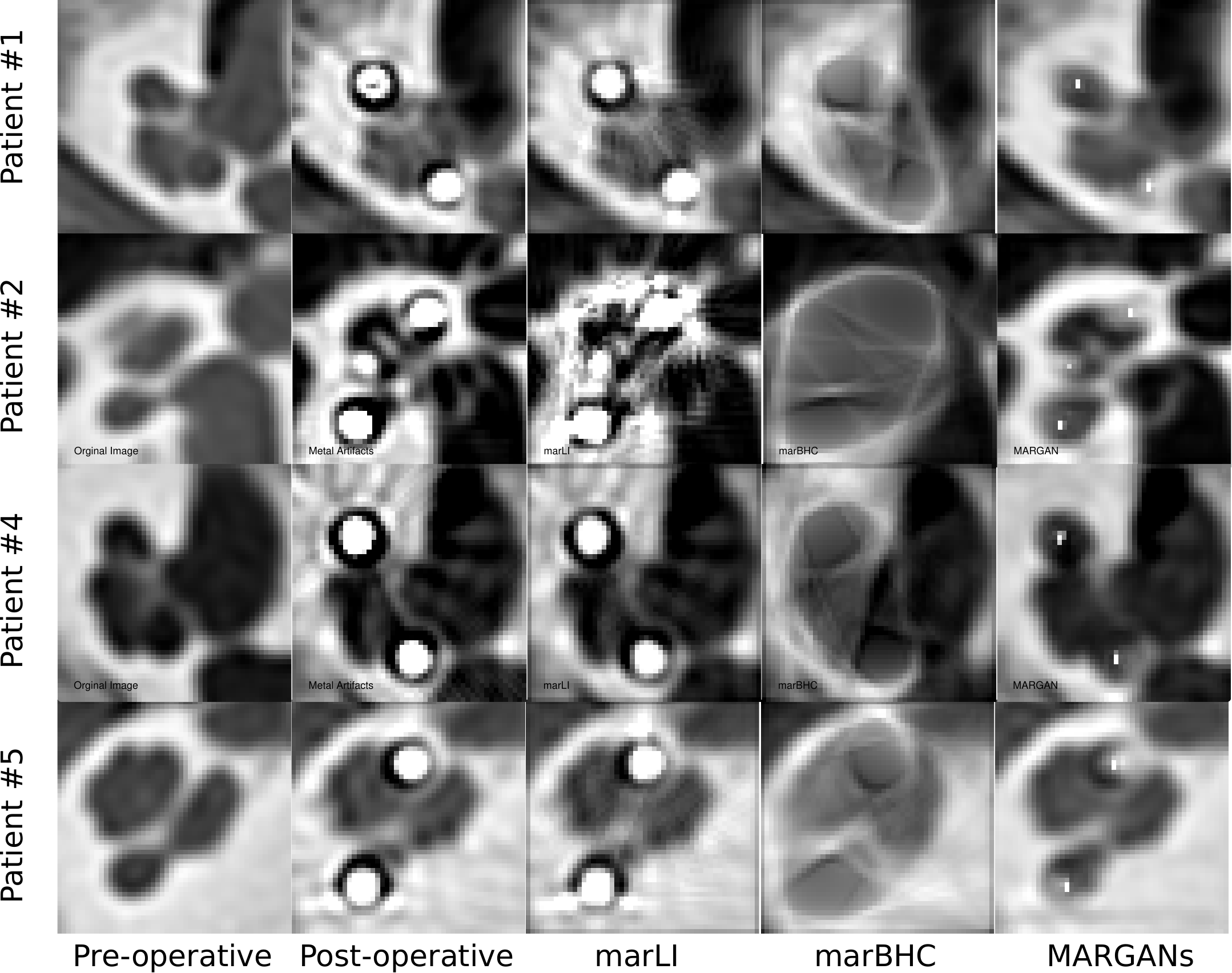}
\caption{Metal {artifact} reduction visualization of MARGAN in comparison with other approaches for patients \#1 - 5.}
\label{fig:comparison}
\end{figure}
\begin{figure*}[!htbp]
\centering
  {\includegraphics[width=2\columnwidth]{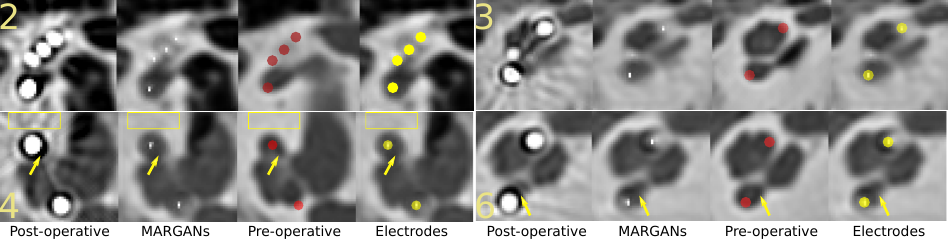}}
  \caption{Results from patients $\#2$ ({top left}), $\#3$ ({top right}), $\#4$ ({bottom left}) and $\#6$ ({bottom right}) for two middle slices (first and second rows). The four columns correspond{ to}: original {postoperative} images, output of MARGANs, registered {preoperative} images with manually positioned electrodes in red and {postoperative} images with electrodes appearing in yellow.}
  \label{fig:evaluationResultPatient6and4}
\end{figure*}

\paragraph{Qualitative Study}
Fig.~\ref{fig:evaluationResultPatient6and4} shows the output of the MARGAN network for four patients on two selected slices together with pre{-} and {postoperative} CT images. The streak {artifact} patterns were largely suppressed by the MARGAN algorithm.  As shown inside the yellow boxes, the artifact patterns were significantly reduced compared to {postoperative}  images.
The cochlear structures that were slightly distorted by the artifacts ({indicated by} yellow arrows) {were} mostly recovered in comparison to the {preoperative image} slices. Finally, the  MARGAN{-}generated images include by design, high intensity pixels at the potential locations of electrode centers. 
The yellow circles are clearly positioned  in the centers of the electrodes  and can help otologists visualize the relative positions of electrodes with respect to the scala tympani. 
\begin{figure}[!h]
\centering
\includegraphics[width=\columnwidth]{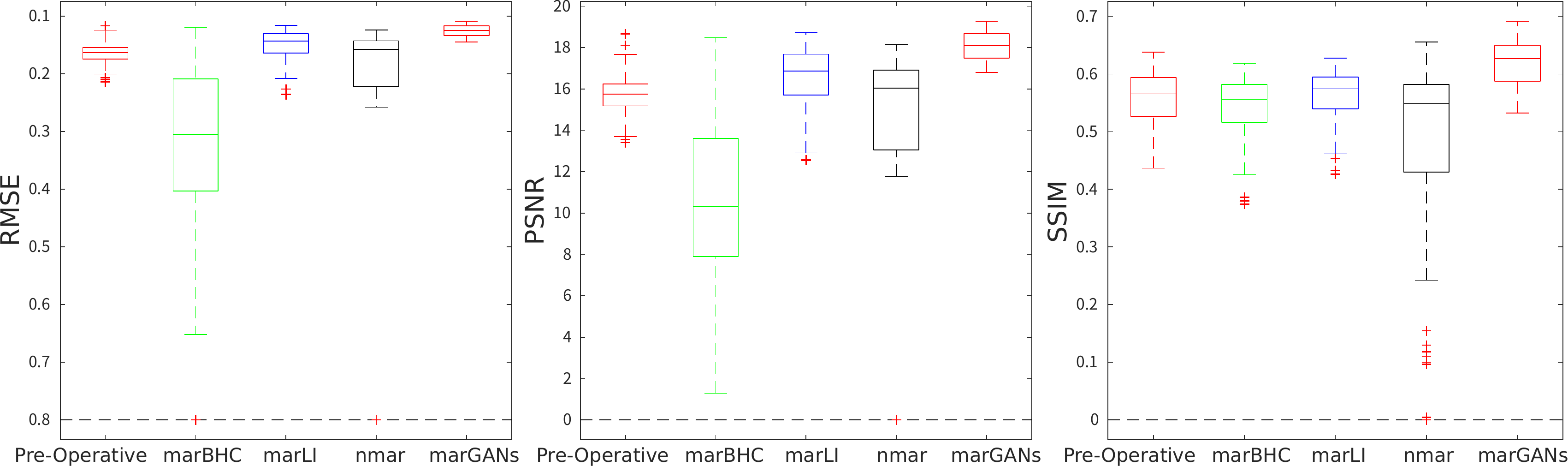}
\caption{The 3D consistency between slices from patient \#1 for three different metrics. We see the MARGAN {algorithm} achieves the best slice consistency in comparison to other approaches.}
\label{fig:errorBar}
\end{figure}
\begin{table}
\centering
\caption{Quantitative evaluation of the MARGAN {approach compared to} marBHC, marLI and Nmar. It shows the performance gain of MARGAN compared to other methods.} %CB clarify this caption - "shows the advance of"? SSIM=STD? Paragraph on "Quantitative Comparison with other MAR algorithms" explains metrics.
\begin{tabular}{cccccc}
{Metric} &{Preoperative}& marBHC      & marLI & Nmar   & MARGAN      \\ 
\hline
PSNR&16.33 &11.59 &16.53 &13.58 & \textbf{18.31} \\
RMSE&0.15      &   0.28       &  0.15&  0.52      &   \textbf{0.12}\\
SSIM&0.58      &    0.56   &       0.55      &    0.52     &     \textbf{0.64}\\
\hline                   
\end{tabular}
\label{tab:evaluation}
\end{table}
\paragraph{Quantitative Comparison with other MAR algorithms} Similar to~\citep{Zhang}, we compared our approach with three open{ }source MAR algorithms{:} MAR with projection linear interpolated replacement (marLI)~\citep{KalenderW}, beam hardening correction (marBHC)~\citep{Verburg} and NMAR \citep{MeyerE}. The visual assessment of the different MAR algorithms is {shown} in Fig.~\ref{fig:comparison}. The MARGAN {approach} clearly outperforms the other three MAR methods in its ability to decrease the texture changes of artifacts and to generate an image similar to the {preoperative} image.  All 33 {postoperative} images were processed by marLI, marBHC, and NMAR. Three global similarity indices,  the {root mean square error} (RMSE), {structural similarity index} (SSIM) and  {peak signal to noise ratio} (PSNR), were computed between the {preoperative} images and the  MAR images generated by {the} three {comparison}  methods and our proposed approach. {These} three  indices {are} reported in Table~\ref{tab:evaluation}{ and} capture  the preservation of visible structures, the errors and {the} quality of the reconstructed images.  Our  method outperforms {the other} MAR  methods for all three metrics (lowest RMSE and largest SSIM and PSNR). In Fig.~\ref{fig:errorBar}, the same indices were computed for all patient $\#1$ image slices to evaluate the spatial consistency of the reconstruction. Clearly the MARGAN approach
exhibits the best {performance,} with a lower mean value and much lower variance. This can be explained by the fact that it is the only MAR algorithm working directly on 3D images. 

\subsection{Impacts of methodological contributions}

We assess the importance of our methodological contributions by evaluating their impact on the generated MARGAN images when they are removed from the computational pipeline. More precisely, we consider {the following} two contributions:
\begin{itemize}
    \item {\em Retinex Loss} When  zeroing the Retinex scale factor $\alpha = 0$ (instead of setting $\alpha =\expnumber{5}{-5}$) during the MARGAN  training, only the $L_{mse}$ loss term is used, which is equivalent to minimizing the $L2$ norm between {the} generated and ground truth {images}. We also { } include in the ablation study the replacement of $L_{mse}$ with the $L1$ norm involving $|I^{train} - G_{w_g}(I^m)|$ terms. 
    %and compare it with the regular case for which .
    \item {\em  Simulation of scatter and electronic noise in artifact simulation} We simulated the image training set with only the beam hardening effect (as in~\citep{Zihao}) or with the full pipeline as described in section~\ref{subsec;simulation}.
\end{itemize}

\begin{table}[!htbp]
\centering
\caption{Ablation Study of Retinex and Physical Simulation}
\begin{tabular}{llll}
Dataset     & PSNR & RMSE & SSIM \\ 
\hline
MARGAN L1 Scatter& 16.67 & 0.1490              & 0.56                      \\
MARGAN L2 Scatter& 18.17 & 0.1257              & \textbf{0.64}                      \\
MARGAN L2+Retinex No-Scatter & 18.02  &   0.1277  & 0.63                \\
MARGAN L2+Retinex Scatter& \textbf{18.31}   & \textbf{0.1242} & \textbf{0.64}         \\
\hline
\end{tabular}
\label{tab:ablation}
\end{table}

In Table~\ref{tab:ablation}, we used the three similarity measures PSNR, RMSE{ }and  SSIM with respect to {the preoperative} images as a way to quantify the impact of those contributions. 

\begin{figure}[htbp]
\centering
  {\includegraphics[width = \columnwidth]{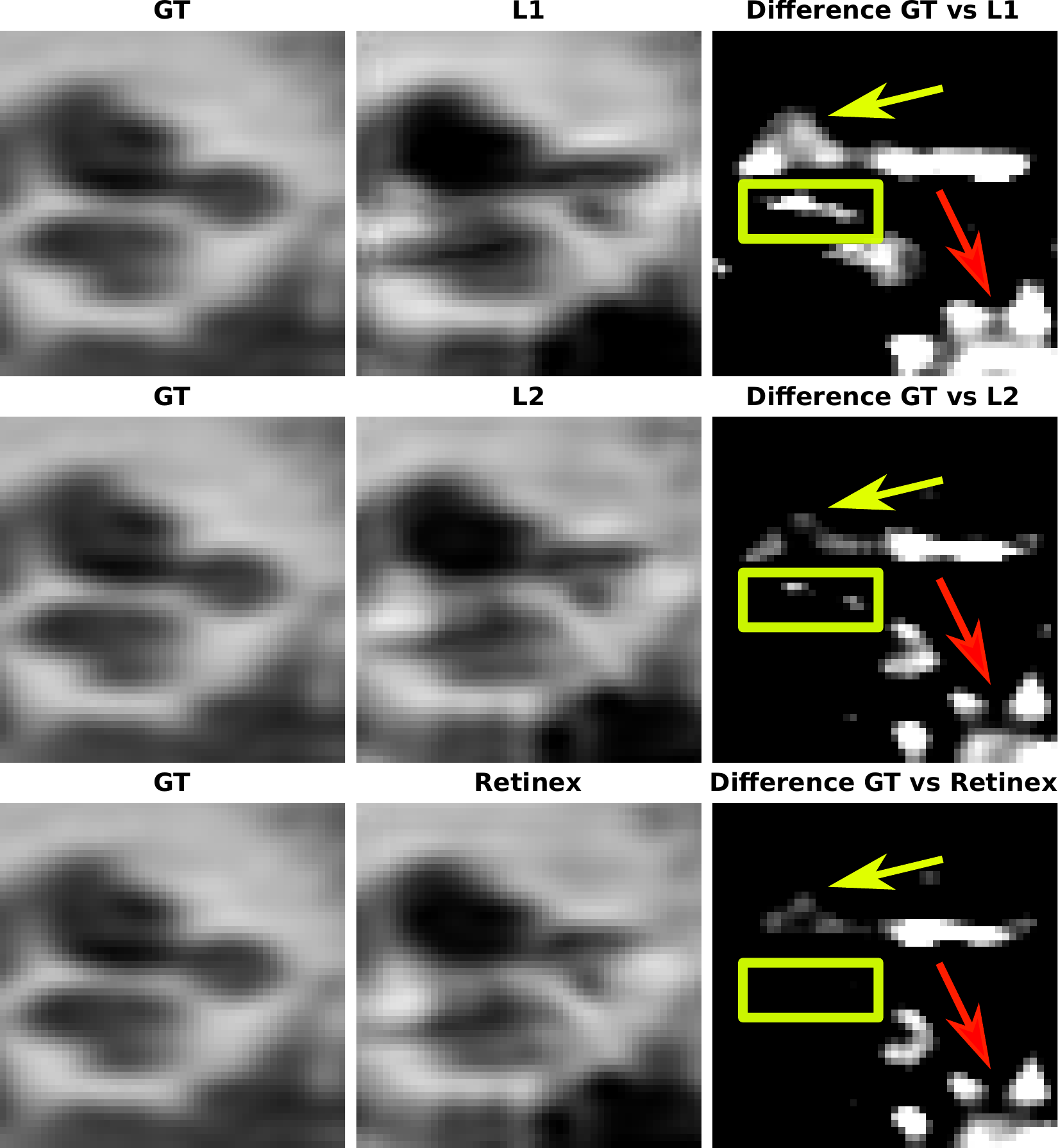}}
  \caption{Qualitative ablation study of Retinex loss effectiveness. The first column is a middle slice of patient \#1, the second column is the {corresponding} outputs of MARGAN with different loss functions and the last column shows subtraction maps between the {first} two columns.}
  \label{fig:ablation}
\end{figure}

{Table~\ref{tab:ablation}} shows that both the addition of scatter and electronic noise in the simulation and the addition of the {Retinex} loss can improve the performance of MAR for all three different metrics. We also see that using a single $L1$ loss function performs worse than the proposed loss combination {approach}. %CB check okay to remove "for MAR task" here, wasn't clear 
A visualization of the image difference output {obtained} using different training loss {functions} is shown in Fig. 
\ref{fig:ablation} with{ }subtraction {maps} between different {output images} and {the} ground truth image. We see from the yellow and red marks in those subtraction maps{ }the effectiveness of the {proposed Retinex} loss function in comparison with using pure L1 and L2 losses.

\subsection{Out-of-sample Test}
To assess the generalization ability of this MARGAN approach,
we explore its performance  on 8 {postoperative} CBCT images{, noting that} the network was trained on CT images. 

\begin{figure*}[h!]
  \centering
  \includegraphics[width=2\columnwidth]{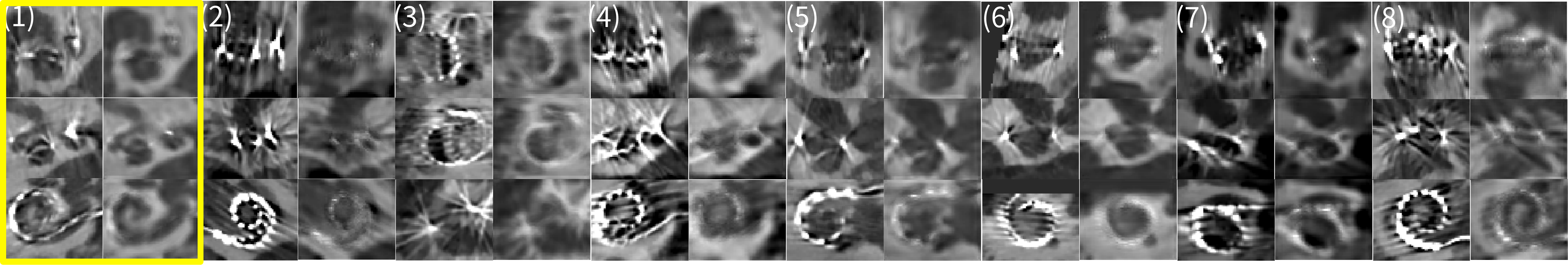}
  \caption{Performance of MARGAN on 8 CBCT {postoperative} images. {The yellow} box shows  three views of {postoperative} images and MARGAN{-}processed images for patient \#1.}
  \label{fig:cbct}
\end{figure*}

In Fig.\ref{fig:cbct}, we see that metal artifacts in CBCT are more extensive and complex than in CT images. Yet, the MARGAN can cope well with those CBCT images and is able to recover most of the cochlear structures.  
%which are difficult for evaluation in artifacts images.

\subsection{CI {Electrode} Position Prediction}
\begin{figure}[!htbp]
\centering
\includegraphics[width = \columnwidth]{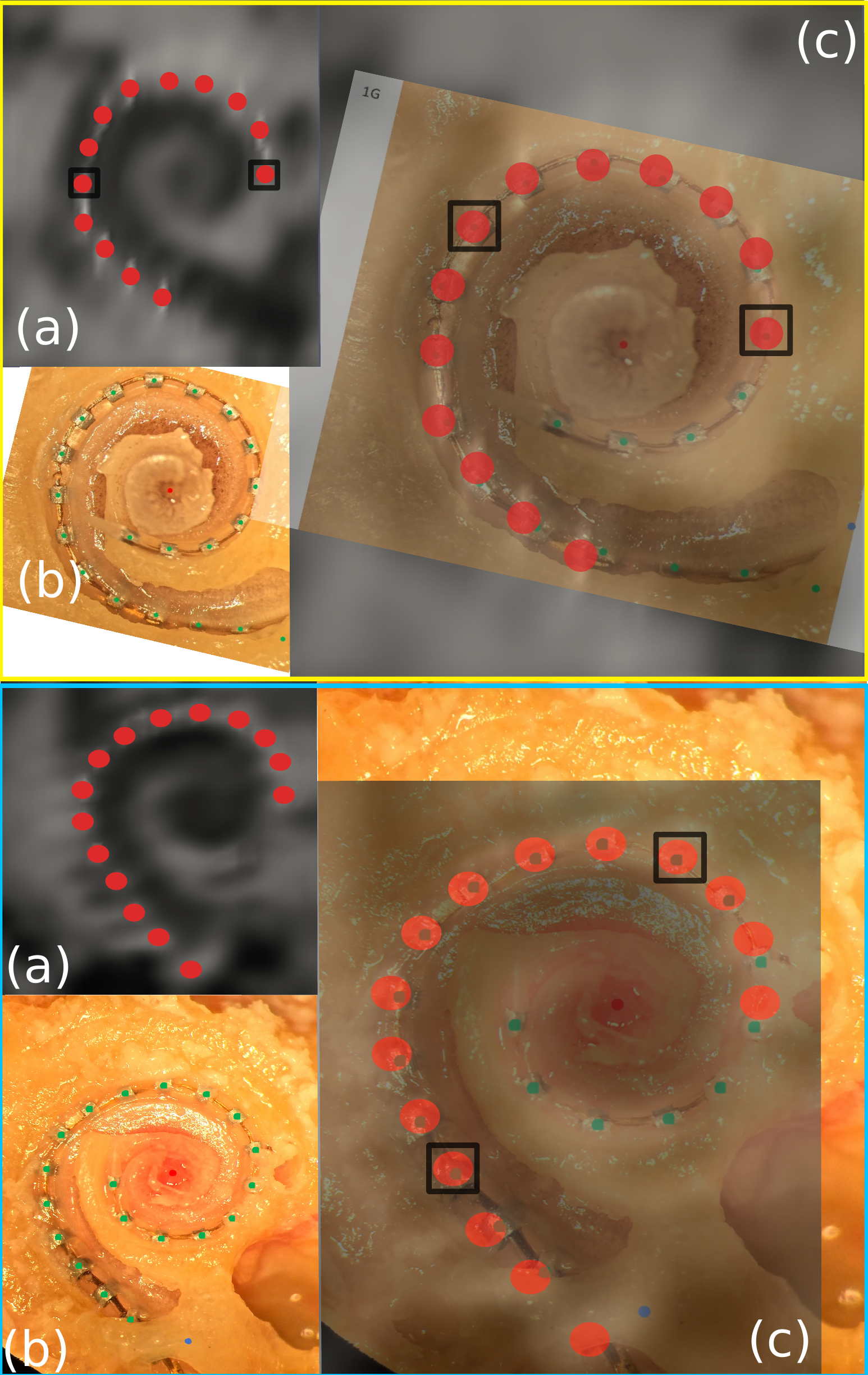}
\caption{Evaluation of the {electrode} position after the application of the MARGAN algorithm on 2 subjects ({top and bottom}); (a) Reformat of a 3D MARGAN image along a plane orthogonal to the modiolar axis. Red circles were manually added at {the} location of high intensity voxels; (b) Image of the cochlea with electrodes inserted after dissection and grinding of the temporal bone; (c) fusion of{ } images (a) and (b) after an affine transform based on the manual correspondence of {the} centers of the two circles outlined by {black squares}. A good overlap of green and red squares is observed.}
\label{fig:electrodeCompare}
\end{figure}

The positioning of CI electrodes in {postoperative}  imaging provides{ }important  information for establishing {a} hearing  prognosis~\citep{Maria,Ingo}  and can be used to improve the cochlear implant programming strategy~\citep{pmid25402603}. The proposed MARGAN algorithm output images where the electrode centers are outlined by voxels in hypersignal as %dots can not only eliminate metal artifacts but also provide the electrodes positions 
shown in Fig.~\ref{fig:evaluationResultPatient6and4} and \ref{fig:comparison}. 
To qualitatively evaluate the {positional} accuracy of those electrode centers in generated MARGAN images, we use pictures of the cochlea acquired after the dissection and grinding of post-mortem temporal bones following CI surgery ( see Fig.~\ref{fig:electrodeCompare}(b)). 
On each generated MARGAN image, a slice having roughly the same position and orientation  as the dissection picture has been manually extracted (Fig.~\ref{fig:electrodeCompare}(a)). 
 Semi-transparent red  circles have been manually positioned on the MARGAN slice at  high intensity voxels while green dots have been positioned by an otorhinolaryngologist on the electrodes visible in dissection pictures. Furthermore, those {two} images have been  registered with an affine transform estimated after selecting {two} corresponding electrodes. The two registered images are fused in Fig.~\ref{fig:electrodeCompare}(c) thus showing the good overlap between green and red circles. This experiment shows that{ }information about the position of the electrodes {causing} the artifacts {was} kept after the application of the MARGAN algorithm.

\section{DISCUSSION}
\label{DISCUSSION}
%\vspace{-2cm}}
Our MARGAN approach combines an artifact simulation pipeline with a 3D GAN network that generates {\em augmented} {preoperative} images from {postoperative} images. The artifact generation algorithm {relies} on three physical phenomena: beam hardening, scatter and electronic noise.  The scatter and noise effects clearly have  less  impact 
on the output image compared to beam hardening.
Yet, these effects were shown in Table~\ref{tab:ablation} to improve the realism of the output of MARGAN when compared to {preoperative} images. The simulation pipeline could {easily be} refined in many ways{, for example,} using {a} more hardware{-}specific energy spectrum, increasing the number of sample energies in the approximation, {or including} more application{-}dependent scatter to primary ratios. This approach could {also} be extended to other imaging systems{,} such as cone beam CT,  dual energy CT or trimodal low-dose X-ray tomography~\citep{synchrotron}. 
%Yet, we have shown those effects are of second
The use of 3D GANs allowed us to generate MAR images with spatial coherence across neighboring slices{,} which is not guaranteed when using 2D slice-by-slice 
MAR methods. Furthermore{, Retinex} loss was introduced to improve the sharpness of the MAR images.  
We show in Tab. \ref{tab:ablation} that the {Retinex} loss can
improve the performance of the MARGAN with a scale coefficient $\alpha =\expnumber{5}{-5}$. 
{However}, an inappropriate $\alpha$ value can introduce distortions in  the MARGAN output. Furthermore, the influence of other hyperparameters in the simulation pipeline on the artifact reduction needs to be further investigated. 

The  MARGAN approach is both data driven (for  the generation of MAR images) and  model driven (for the generation of training image pairs). 
This is in contrast to purely data-driven MAR methods {that} either {rely} on  pairs of {pre-} and {postoperative} images~\citep{Jia} or on non-paired data~\citep{Nakao}. The collection of image pairs, with and without artifacts{,} is mostly restricted to images acquired before and after an intervention {such as} CI  insertion. The use of such pairs makes the 3D GAN  fairly effective at removing artifacts in {postoperative} images. However, the collection of those images may be difficult and an intra-patient image registration is required. The MARGAN pipeline aims at reaching the same efficiency but by replacing the {postoperative} image with a simulated one. This makes the MARGAN algorithm applicable to a larger set of clinical cases where such image pairs cannot be gathered{,} for instance in the case of hip, shoulder or knee {prostheses}. The use of CycleGAN on non-paired images as in \citep{Nakao} is very appealing{, because} it avoids both artifact simulation and collection of paired images. However, it has{ }only been tested to remove large artifacts{,} such as those caused by dental {fillings,} and with limited quantitative assessment. 

Another advantage of the artifact simulation approach in MARGAN  is its ability to augment the generated MAR image with voxels indicating the location of the metal part. In the case of CI {postoperative} images, it enables{ }visualization in the same image of both the cochlea and the implant electrode centers. Note that the augmentation of the MARGAN image is only 
optional in this framework, {because} the metal{-}free image can replace the augmented image as $I^{\mathrm{train}}$ in the loss function of the 3D GAN.

Specifically, in {the} cochlear metal {artifact} reduction problem, we see from Fig.~\ref{fig:comparison} and Tab.~\ref{tab:evaluation} that almost all the traditional MAR approaches have degradation problems in terms of reconstruction image quality. 
{It} was reported in \citep{MeyerBergner} {and} \citep{Diehn} {that} sinogram {inpainting-based} methods can introduce new artifacts. {These} artifacts can have a severe impact on image quality if the metallic parts and artifacts occupy a large area of space {in} the image{,} which is typically the {case} for {the} CI electrodes {discussed} here. However, the risk of quality degradation
is not applicable for MARGAN as the image domain{-}based methods do not need to access the sinogram and {the} Radon transform.

A limitation of MARGAN lies in the relative complexity of implementing the simulation of metal artifacts in CT images. This is especially true for the scatter effect,{ }which only adds {a} marginal{ }gain in {realism to} the generated images. A more thorough study should be performed to evaluate the  level of realism required in the simulation pipeline to improve the MARGAN output. 
Simulating the insertion of metal parts can also be complex as it requires a segmentation algorithm to locate the region of insertion. But this complexity is rewarded by the ability to generate a vast training set accounting for variations in patient anatomy or implant design.  

Learning{-}based MAR methods {were} shown to outperform traditional sinogram-based MAR approaches in several {previous} works~\citep{Jianing-b, Zhang, Zihao}. But by design, the performance of those supervised methods {depends} on the chosen training set and {they} are application{-}specific algorithms. Their integration into a clinical workflow remains to be demonstrated, in particular due to their potential lack of robustness. The successful application of MARGAN on CBCT images unseen during  training is an encouraging sign of the  generalization ability of MARGAN{, though} further studies are required. 

Finally a limitation common to all MAR methods is the difficulty {of evaluating performances} quantitatively{, }due to the lack of ground truth data. The use of paired {pre- and postoperative} image data {enables} quantitative comparison through global similarity indices ({such} as PNSR, RMSE) but {is} also dependent on the registration quality of the two images. Images with synthetic artifacts created by image processing were also considered in \citep{Nakao}{,} for instance{,} but they are computationally intensive to reach sufficient realism. Physical anthropomorphic phantoms are a useful alternative for MAR assessment~\citep{BolstadPhantom} but are limited by the number of phantoms considered.

\section{CONCLUSION}
\label{CONCLUSIONS}
In this paper, we have introduced a simulation{-}based 3D GAN to attenuate metal artifacts in CT images. The network is trained on a thousand regular CT images without any artifacts and their  corresponding images where metal artifacts have been simulated. We have demonstrated the introduction of scatter and electronic noise effects in addition to beam hardening in an efficient computational pipeline. The complexity of scatter simulation has been alleviated by {precomputing}
the impact of scatter on a generic head phantom where metal parts have been introduced. A {Retinex} loss was introduced to enhance visible edges  in the generated images. The MARGAN approach was evaluated on CT and CBCT images of the inner ear with cochlear implants { }inserted. The proposed approach provided images close to {preoperative} images and {outperformed} open source MAR methods. Furthermore, images generated by MARGAN {included} the location of the electrode centers{,}  which is useful for assessing the quality of implant surgery. 

The trade-off between the complexity of artifact simulation and MARGAN output requires additional study, and we will also investigate the impact of MARGAN images {on}
the automatic registration of {pre-} and {postoperative} images.
 
% \section*{ACKNOWLEDGMENT}
% This work was partially funded by the regional council of
% Provence Alpes Côte d’Azur, by the French government through the UCA JEDI
% ”Investments in the Future” project managed by the National Research Agency
% (ANR) with the reference number ANR-15-IDEX-01, and was supported by the
% grant AAP Santé 06 2017-260 DGA-DSH.
 
\printcredits

\bibliographystyle{cas-model2-names}

\bibliography{Bibliography}


% \bio{figs/pic1}
% Author biography with author photo.
% Author biography.

\end{document}